\documentclass{article}
\usepackage{spconf,amsmath,graphicx}

\usepackage{booktabs}
\usepackage{multirow}
\usepackage{amssymb}
\usepackage{booktabs}
\usepackage{array}
\usepackage{arydshln}
\usepackage{pifont}
\usepackage{color}
\usepackage{colortbl}
\usepackage[x11names]{xcolor}
\usepackage[breaklinks,colorlinks]{hyperref}

\usepackage[capitalize]{cleveref}
\crefname{section}{Sec.}{Secs.}
\Crefname{section}{Section}{Sections}
\Crefname{table}{Table}{Tables}
\crefname{table}{Tab.}{Tabs.}

\usepackage{fancyhdr}
\fancypagestyle{firstpage}{
  \fancyhf{}
  \fancyhead[C]{Preprint paper version accepted at 2026 IEEE International Conference on Image Processing (ICIP)} % set the header
}

\usepackage[OT1]{fontenc}

\begin{document}

\title{Improving Viewpoint-Invariance and Temporal\\ Consistency for Action Detection}

\name{Yannick Porto$^{\star}$ $^{\dagger}$ \qquad
Renato Martins$^{\star}$ \qquad
Thomas Chalumeau$^{\dagger}$\qquad
Cedric Demonceaux$^{\star}$
}

\address{$^{\star}$ Université  Bourgogne Europe, CNRS, ICB \\
  $^{\dagger}$ TEB Group, Prynel SAS
  }

\maketitle
\begin{abstract}

Viewpoint change invariance and action temporal consistency are critical aspects for the effective deployment of human action detection of untrimmed videos. Existing appearance-based video detection methods often struggle with limited viewpoint diversity during training, while motion-based detection approaches frequently fail to model fine-grained temporal relationships across consecutive motion windows. This paper introduces a novel two-stage action detection approach designed to improve both view-invariance and global temporal coherence properties. In the first stage, we extract motion features from augmented virtual viewpoints, solely used at training. Then, the second stage introduces a new view-invariant, multi-scale temporal encoder based on selective state-space sequence modelling to aggregate information across viewpoints and time scales. Experiments on PKU-MMD and BABEL benchmarks demonstrate that this approach significantly outperforms state-of-the-art methods in all considered splits. Code and trained models are available at: \url{https://icb-vision-ai.github.io/HydraView-TAD}
\end{abstract}

\begin{keywords}
Action Detection, Human Motion Analysis, Action Recognition.
\end{keywords}

\thispagestyle{firstpage}

\vspace{-0.3cm}

\section{Introduction}

Temporal Action Detection (TAD) aims at recognizing and localizing human actions in long, untrimmed video sequences. Unlike trimmed action recognition, TAD requires not only identifying the action category but also accurately determining its temporal boundaries, making it a fundamental yet challenging problem for activity understanding. {A critical aspect often overlooked in video-based methods is of explicitly considering {viewpoint variability} conditions. Changes in viewpoint can significantly alter the perceived action patterns, leading to inconsistencies in action representations and degraded detection performance. Beyond viewpoint changes, actions also typically exhibit strong {temporal consistency} across adjacent frames, where motion evolves smoothly over time. Local transitions provide essential cues for accurate boundary localization, and remains insufficiently addressed by existing methods treating motion data. As a result, learning view-invariant representations that preserve temporal coherence across frames is essential for reliable TAD.}
\begin{figure}[t]
\centering
\includegraphics[width=\linewidth]{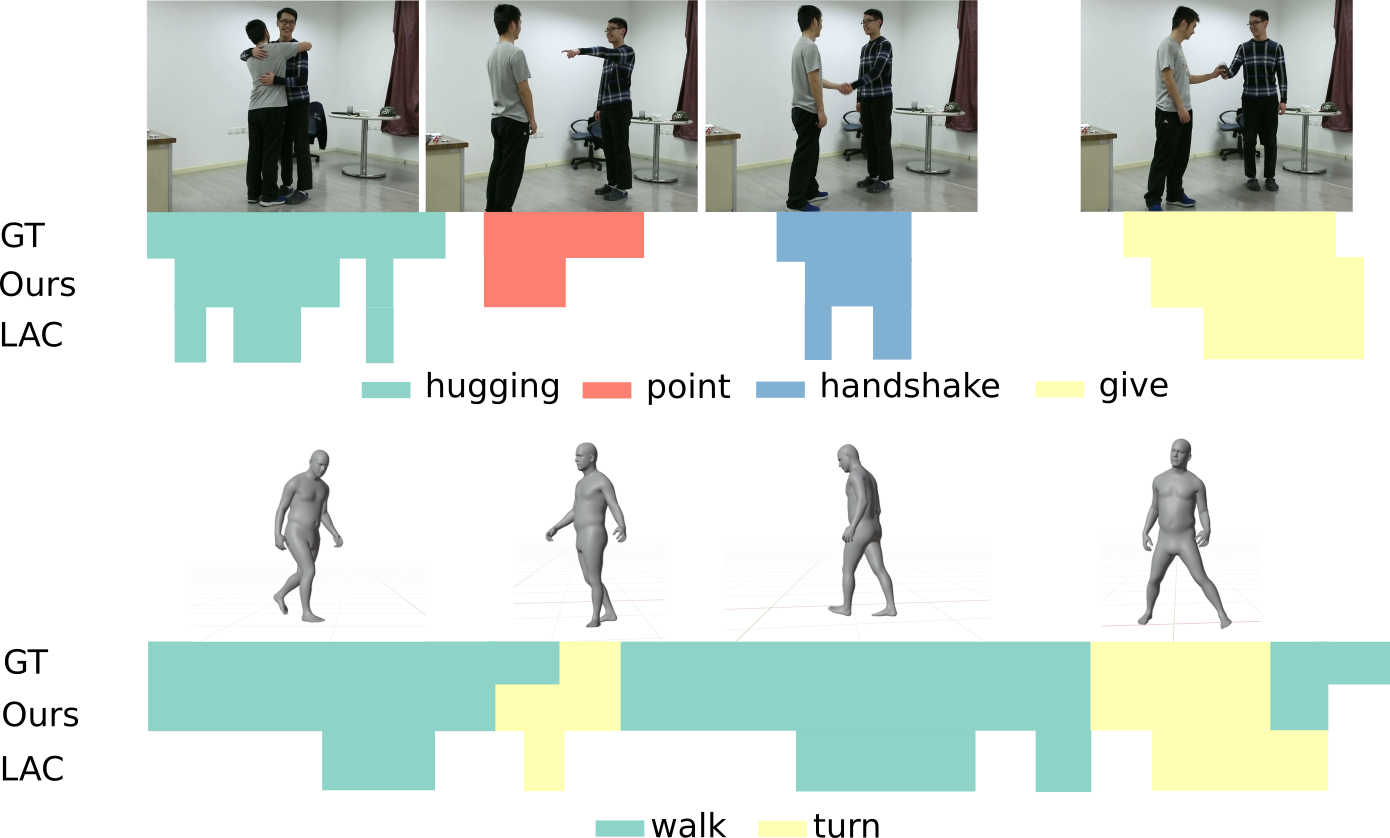}
\caption{Action detection on two sequences from PKU-MMD and BABEL datasets. GT indicates ground truth action labels per frame. Please notice the improved detection results from the proposed model (Ours) when compared to the baseline LAC for the actions ``pointing'' and ``walk''.}
\label{fig:qualit_examples}
\end{figure}

In this context, this paper proposes a novel multi-view TAD framework that jointly addresses view-invariance and temporal modeling in single-view sequences at inference. Our method decomposes untrimmed videos into small temporal windows and processes them in two main stages. First, we introduce a multi-view learning for window encoding with a new spatio-temporal encoder, suitable to reinforce temporal coherence in small periods of time. Compared to other skeleton-based action detectors~\cite{yang2023lac,tian2025duoclr,gokay2025skeleton}, the spatial features are then frozen to train the temporal encoder. Learning relations on windows is costly, especially for long sequences, and moreover for multiple views. We therefore operate the same strategy than RGB-based TAD methods~\cite{dai2021pdan,zhu2024dual,sinha2025ms} with a two-step training.

\begin{figure*}[t]
\centering
\includegraphics[width=\linewidth]{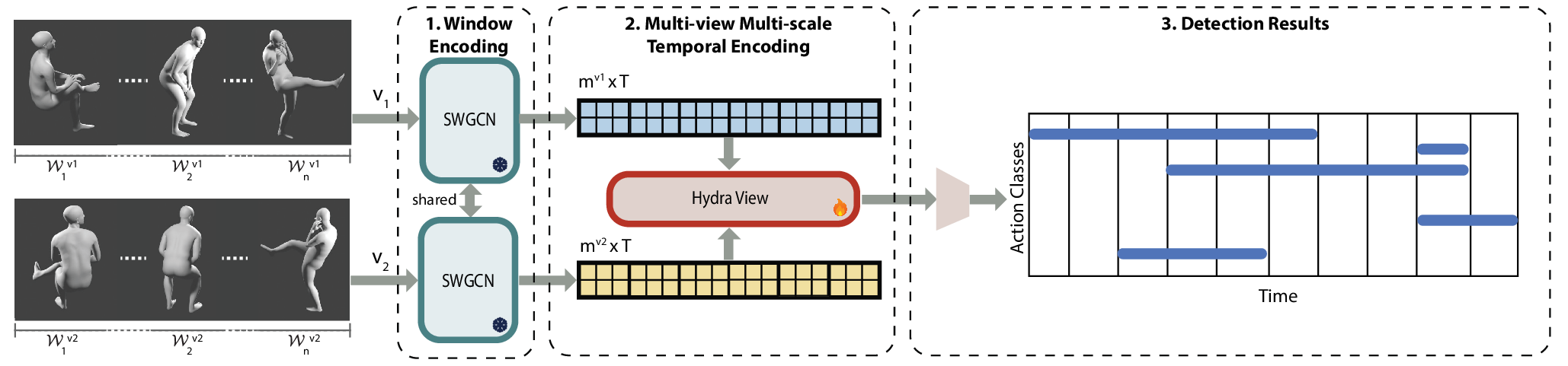}
\caption{Overview of our temporal action detection method with two viewpoints. For each input video viewpoint, an untrimmed sequence is encoded with a spatio-temporal encoder to generate features with improved view invariance. These features are then refined by our multi-view and multi-scale temporal encoder (HydraView) for localizing each action over time.}
\label{fig:overview}
\end{figure*} 

Second, we propose {a view-invariant} multi-scale temporal encoder, termed HydraView, which aggregates information across both time and views to perform frame-level action detection, as illustrated in \cref{fig:qualit_examples}. HydraView is built upon a sequence of ViewMamba blocks that extend recent state space model architectures to consider temporal and {multi-view settings}. Each block captures temporal relations at a specific scale while selectively scanning across both temporal and viewpoint dimensions. This design enables the model to handle simultaneously different viewpoints of the video at training and constrains the model to learn a similar action feature representation, reinforcing viewpoint-invariance. Then, a dedicated multi-scale fusion mechanism integrates these representations to produce accurate per-frame, multi-label action predictions.

{Our method requires a set of different viewpoints solely at training time, and it is capable of considering at inference both single view and multi-view settings. Yet, to allow a fair comparison to competitors all results are shown in the context of a single viewpoint action detection. The main contributions of this paper are twofold:
\emph{i)} {To the best of the authors' knowledge, this is the first action detection model for learning temporal consistency on motion data captured across different viewpoints for the task of human temporal action detection. To achieve this, we transpose a two-stage training strategy commonly adopted in video-based action detection, enhanced with a multi-view motion encoding strategy, which is done solely at training time.}
\emph{ii)} We introduce a new multi-view, multi-scale temporal architecture based on state space sequence modelling that jointly reasons over temporal dynamics and viewpoint variations with linear complexity in sequence length. The experiments show the capabilities of the designed approach, outperforming recent video and motion state-of-the-art temporal action detection methods on both PKU-MMD and BABEL benchmarks.

\vspace{-0.3cm}

\section{Related Work}

\vspace{-0.2cm}

\subsection{Video-based Action Detection}

Early approaches to temporal action detection were largely proposal-based, drawing inspiration from object detection to generate candidate temporal segments. Although effective for sparsely annotated videos, these methods were computationally expensive and poorly suited for dense per-frame predictions. To address these limitations, temporal convolution-based models were introduced to efficiently process long videos~\cite{dai2021pdan,dai2022ms,Dai:TPAMI:2023}. Such approaches, including Temporal Convolutional Networks, employ hierarchical temporal convolutions for fine-grained action detection~\cite{dai2022ms}, but their inherently local receptive fields restrict the modeling of long-range temporal dependencies~\cite{wang2022empirical}. More recently, transformer-based architectures have demonstrated strong performance in capturing both short- and long-term temporal dependencies~\cite{zhu2024dual}, albeit at high computational cost due to quadratic attention complexity. To improve scalability, state space sequence models (SSMs) have emerged as efficient alternatives with linear complexity for long-sequence modeling~\cite{gu2024mamba,zhu2024vision,liu2024vmamba,lu2025jamma}. Several recent TAD methods integrate SSMs to balance expressiveness and efficiency~\cite{liu2024harnessing,chen2024video,sinha2025ms}. In particular, CausalTAD~\cite{liu2024harnessing}, TimeMamba~\cite{chen2024video}, and MS-Temba~\cite{sinha2025ms} explore different SSM-based designs for temporal action detection. Our approach builds upon these recent advances by leveraging multi-scale SSM operators, but considering a {multi-viewpoint action recognition scheme}.

\vspace{-0.3cm}

\subsection{Motion-based Action Detection}

Several recent motion-based action detection approaches exploit auxiliary learning schemes to encode discriminative motion cues~\cite{yang2023lac,tian2025duoclr}, but they lack from temporal information since temporal features are globally pooled from all the sequence. Following this trend, ~\cite{yang2023lac,tian2025duoclr,gokay2025skeleton,wengUSDRL25} focus on learning view-invariant skeleton representations at the frame level by training motion encoders on short temporal windows observed from varying viewpoints. {Similarly to these works, we design a motion encoder to learn view-invariant representations with skeleton sequences observed from virtual viewpoints.} These methods also refine per-frame features, but do not consider the relations between the different adjacent windows. {In contrast, our proposed approach explicitly models local and long-range temporal dependencies at multiple scales, following strategies commonly adopted in video-based action features}~\cite{dai2021pdan,dai2022ms,tian2025stitch}.

\section{Method}

This section presents the designed temporal action detection method to encapsulate properties from multiple viewpoints and long temporal sequences, as illustrated in \cref{fig:overview}. Following recent TAD methodologies~\cite{dai2021pdan,dai2022ms,zhu2024dual}, we start by pre-processing the input video sequence in small windows of time which are then encoded to learn relations along the whole video. Our methodology is composed of two main components: 1) A spatio-temporal motion encoding strategy to encode local windows of actions into spatially meaningful features, and 2) A new multi-view and multi-scale temporal feature encoder which learns the relations between the different windows over time and from the different viewpoints.

\subsection{Window Feature Motion Encoding}

\begin{figure}[t]
\centering
\includegraphics[width=\linewidth]{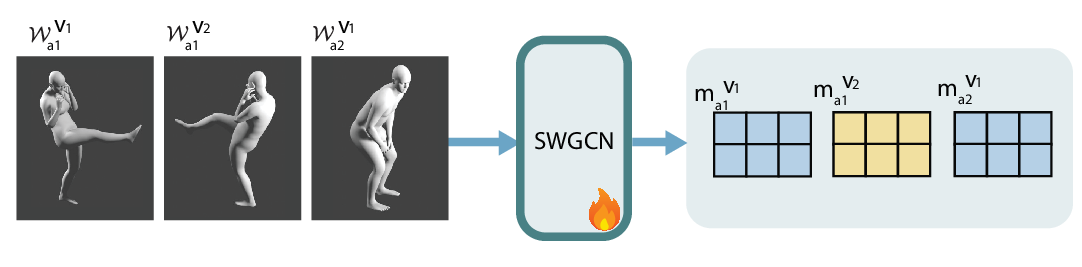}
\caption{Spatial feature encoding. Motion features are learned by small time windows for each viewpoint. The encoder model (SWGCN) is then frozen to extract window features with improved temporal consistency.}
\label{fig:spatial_encoding}
\end{figure} 

During training, we consider as input a 3D motion sequence which are divided into temporal windows. Each window $\mathcal{W}$ in $\mathbb{R}^{F \times J \times 3}$ consists of $F$ frames with $J$ body joints. The first step of our training scheme is to obtain projected 2D skeleton sequences, by simulating $V$ virtual cameras with known parameters. As a result, each temporal window generates $\mathcal{W}^v$ distinct 2D motion sequences inside $V$ views. We simulate occlusions by masking limbs behind the torso joints. These projections provide diverse viewpoints and joint visibility patterns derived from a single underlying 3D motion. Following standard temporal action detection practices, each temporal window is composed of $F=16$ frames. 

\subsubsection{Motion Feature Extraction}

We start adopting a spatio-temporal graph convolution network~\cite{Yan2018SpatialTG} as in ~\cite{Shi2sAGCN,yang2021unik,liu2025revealing} to encode the features for small temporal windows. Spatio-temporal networks are designed for action recognition with larger windows of time (around 100 to 300 frames).
When applied to short temporal windows, these design choices (temporal downsampling, batch normalization, or kernel size)  can lead to a mismatch between the temporal receptive field and the available context, resulting in oversmoothing and reduced sensitivity to fine-grained motion cues. The specific architectural adaptations and tuned parameters used to address this issue are detailed in the supplementary material.
We call this network Spatial-Window Graph Convolution Network (SWGCN) as depicted in \cref{fig:spatial_encoding}.

{Subsequently to refine and improve the temporal consistency of these initial features at window level, we adopt a two-stage encoding strategy, inspired by RGB-based temporal action detection methods. In the first stage, the spatial encoder is trained from scratch using the multiple $\mathcal{W}^v$ observations as input. As illustrated in \cref{fig:spatial_encoding}, a same action $a1$ observed from two different viewpoints $v1$ and $v2$ is learned jointly and distinguished from a different action $a2$. During this stage, each temporal window is supervised with its corresponding action label using a cross-entropy loss. After convergence, the weights of the spatial encoder are frozen and window-level features are extracted into a vector $\mathbf{M}^{vi} = [m_{a1}^{vi}, m_{a2}^{vi}, ...]$ for the $i^{th}$ view (left of \cref{fig:overview}).

\begin{figure}[t]
\centering
\includegraphics[width=\linewidth]{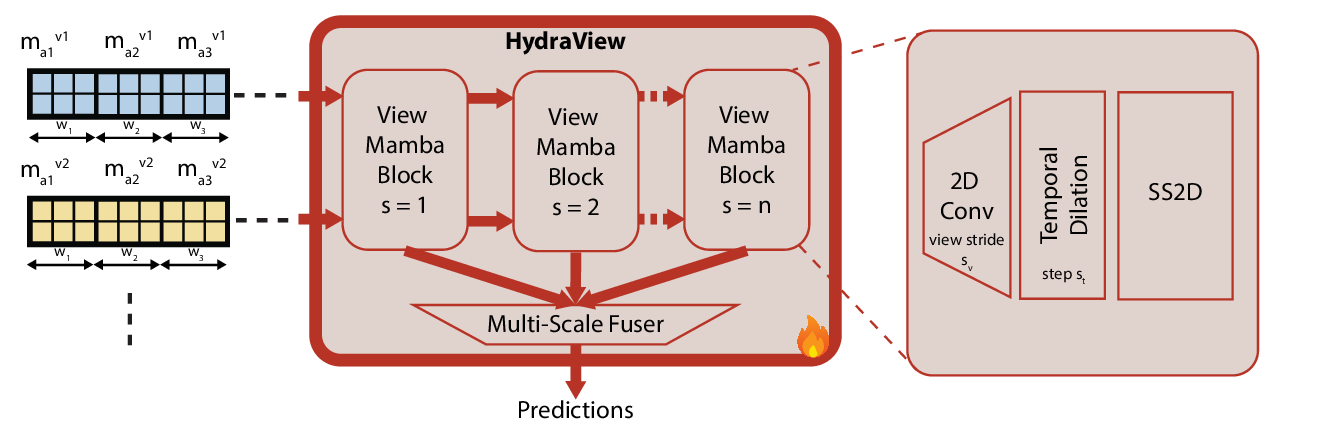}
\caption{Multi-view Multi-Scale Learning for Temporal Encoding with HydraView. HydraView is composed of multiple ViewMamba blocks each responsible of a specific scale in views and time. The ViewMamba block is a 2D convolution followed by a dilation module scaling the input for SS2D. The representation from each ViewMamba block is then processed by a Multi-Scale Fuser to provide the action predictions.}
\label{fig:temporal_encoding}
\end{figure} 

\subsection{Multi-View and Scale Temporal Encoding}
\label{sec:temporal_encoding}
The proposed multiview action detection scheme requires the modelling of relations between the different views in the consecutive frames as well as the temporal consistency of actions in the whole sequence. For that, we propose HydraView (\cref{fig:temporal_encoding}), a new network composed of multiple ViewMamba blocks, responsible of learning a unique representation from the temporal windows coming from all viewpoints. {Building upon~\cite{sinha2025ms}, each ViewMamba block is designed to model temporal dependencies at a specific scale, while explicitly adapting the architecture to the multi-view training regime.}

\subsubsection{Scaling in ViewMamba}
\label{sec:viewmamba}

Each ViewMamba block depited in \cref{fig:temporal_encoding} operates at a specific view–temporal scale. Let $\mathbf{M} \in \mathbb{R}^{V \times T \times C}$ denote the input multiview feature tensor, where $V$, $T$ and $C$ are the numbers of views, temporal steps, and feature dimension respectively. The block begins with a $2$D convolutional projection (\cref{fig:temporal_encoding}) that applies a stride of $s_v$ along the view dimension while preserving the input temporal resolution. This view-strided convolution is defined as
\begin{equation}
\mathbf{M}^{(s_v)}_{k,t}
=
\sum_{i=0}^{K_v-1}
\sum_{j=0}^{K_t-1}
\mathbf{W}_{i,j}
\,
\mathbf{M}_{\,k s_v + i,\; t + j},
\end{equation}
where $\mathbf{W} \in \mathbb{R}^{K_v \times K_t \times C}$ denotes the convolution kernel, $K_v$ and $K_t$ are the kernel sizes in the view and temporal dimensions, and $(k,t)$ are the indexes of the output tokens. Following this convolutional projection, a temporal dilation operation is applied to increase the effective temporal receptive field. In particular, the convolutional responses are sampled with a dilation factor of $s_t$ along the temporal dimension, such that temporal interactions are computed over indices spaced by $s_t$. This dilated temporal processing enables the ViewMamba block to model longer-range dynamics while operating on a reduced set of view-wise tokens.

\subsubsection{Multi-View Temporal Scanning}
Subsequently, the multiview modeling is performed with a selective state space model (SSM) inspired by Mamba and its $2$D extension~\cite{gu2024mamba,liu2024vmamba}. Given an input sequence ${\mathbf{u}}_{t}$, {derived from temporal dilation}, the SSM dynamics are defined as:
\begin{align}
\mathbf{h}_{t} &= \mathbf{A}\mathbf{h}_{t-1} + \mathbf{B}\mathbf{u}_t, \\
\mathbf{y}_t &= \mathbf{C}\mathbf{h}_t + \mathbf{D}\mathbf{u}_t,
\end{align}
where $\mathbf{h}_t$ denotes the hidden state and $\mathbf{y}_t$ the output at position $t$. In Mamba, the state transition parameters are input-dependent and computed through learnable projections:
\begin{align}
\mathbf{B}_t &= \mathbf{W}_B \mathbf{u}_t; ~
\mathbf{C}_t = \mathbf{W}_C \mathbf{u}_t; ~
\boldsymbol{\Delta}t = \text{softplus}(\mathbf{W}\Delta \mathbf{u}_t),
\end{align}
with $\mathbf{A}$ parameterized as a diagonal matrix, $\mathbf{B}$,
$\mathbf{C}$ and $\mathbf{D}$ the weighting parameters discretized using $\boldsymbol{\Delta}_t$~\cite{liu2024vmamba}. This formulation enables efficient long-range dependency modeling while preserving linear-time complexity.

To jointly reason over temporal and view dimensions, the SSM scan is performed along four directions: (i) forward and backward along the view dimension, and (ii) left-to-right and right-to-left along the temporal dimension, as illustrated in \cref{fig:view_scan}. {By introducing this adapted bidirectional 2D scanning, our model effectively retrieve relevant information across viewpoints while maintaining robustness to occlusions or missing detections over time.}

\begin{figure}[t]
\centering
\includegraphics[width=\linewidth]{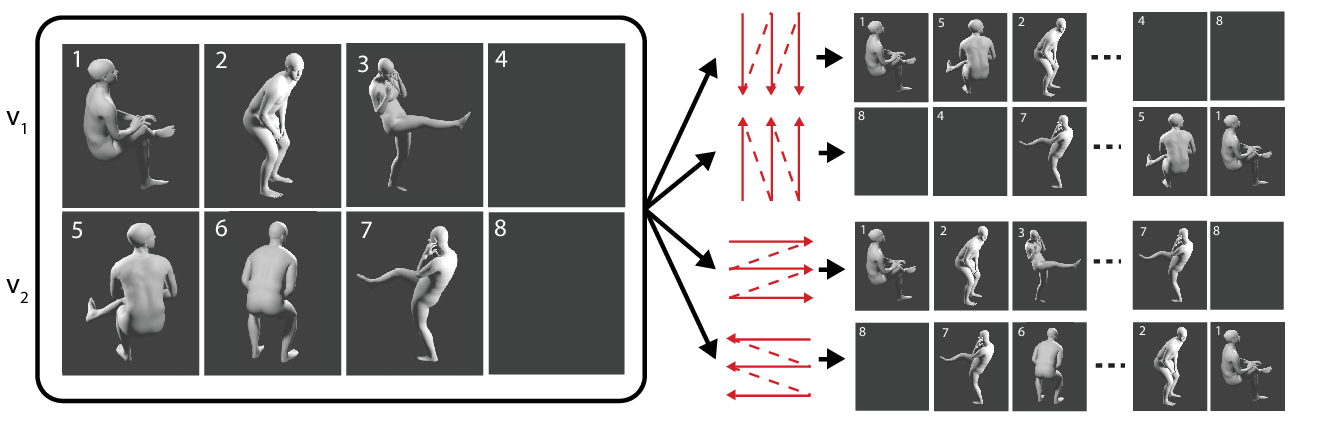}
\caption{Designed bidirectional scanning strategies along both view and temporal dimensions. This scheme shows actions seen from two viewpoints, and final empty cells mean a missing skeleton detection.}
\label{fig:view_scan}
\vspace{-0.3cm}
\end{figure}

We adopt a sequence of three ViewMamba blocks to process the sequence at different temporal scales. Concretely, each $s^{th}$ block parses the sequence with a step of $s$, learning temporal relations between spaced frames. {The different representations are then fused by a Multi-Scale Fuser: a  VisionMamba~\cite{zhu2024vision} component applying a bi-directional horizontal scan to the multi-scale dimension before providing the final predicted actions.} The designed HydraView is the key technical aspect of the method, providing most of the observed performance improvement, and consistently outperforms the state space modeling from MS-TEMBA~\cite{sinha2025ms} (as shown in the study on \cref{tab:ablation}). In multi-label temporal action detection, each temporal segment indexed by $t \in \{1,\dots,T\}$ may contain multiple action classes simultaneously. Therefore, this model is trained with independent binary cross-entropy (BCE) losses  for each considered action classes per time step.

\vspace{-0.2cm}

\section{Experiments}

\noindent \textbf{Experimental Setup.}
We train the motion encoder SWGCN with a feature dimension $d=384$. The HydraView model responsible to enforce temporal coherence and viewpoint change invariance is composed of 3 ViewMamba blocks (each one in one scale), with an output dimension of 192 in each 2D convolution, a view stride of $s_v=2$ and a dilation rate of $s_t=s$ for the $s^{th}$ ViewMamba block. Each ViewMamba has state dimension 16. The number of virtual viewpoints at training was $V=12$, with different body orientations spaced by $30^{\circ}$. The model optimization was done with AdamW, learning rate of $0.00045$, and batch size $4$. All experiments were done on an NVIDIA RTX 5000 Ada GPU.\\

\vspace{-0.3cm}
\noindent \textbf{Datasets.}
We have considered two widely-adopted public available action detection benchmarks (PKU-MMD and BABEL) for the evaluation.
 \textbf{PKU-MMDv1}~\cite{liu2017pku} is an untrimmed video dataset containing 1076 sequences of around 4 minutes each. It has been recorded on three cameras {positioned with a 45° angular separation between each.} We consider the event-based mAP metric~\cite{yang2023lac} for evaluation with a single view at inference as the competitors. {Both evaluated cross-view (CV) and cross-subject (CS) splits involve viewpoint variations.}
\textbf{BABEL}~\cite{BABEL:CVPR:2021} contains 43 hours of 3D human motion sequences from AMASS~\cite{AMASS:ICCV:2019}, annotated with over 63,000 frame-level labels across more than 250 action categories. We follow the exact experimental protocol adopted by (Yu and Fujiwara)~\cite{Yu_Fujiwara_2023} to build three subsets of actions.\\

\vspace{-0.3cm}
\noindent \textbf{Baselines.} We have selected several relevant recent action detection methods for motion and video data, as shown in Tables \ref{tab:pku-mmd-results} and \ref{tab:babel_seg}. These competitors contain the current state-of-the-art approaches on each benchmark: LAC~\cite{yang2023lac}, USDRL~\cite{wengUSDRL25}, ASOT~\cite{xu2024temporally}, SMQ~\cite{gokay2025skeleton} and S-WTAL~\cite{Yu_Fujiwara_2023}.

\subsection{Detection Results}

On PKU-MMD dataset shown in \cref{tab:pku-mmd-results}, our network achieves a mean Average Precision (mAP) exceeding 90\% at IoU thresholds of 10\% on both splits.
%and 25\% IOUs.
Moreover, at an IOU of 50\%, our network obtains a gain of +5.2\% compared to the recent state-of-the-art USDRL~\cite{wengUSDRL25} on this dataset. {These findings can be also noticed in the provided qualitative results shown in \cref{fig:qualit_examples}. The detection improvement in comparison to LAC~\cite{yang2023lac} is notably due to two aspects: actions such as ``pointing'' are better detected due to the view-invariant features. The improved temporal consistency of motion segments is also observed for actions such as ``hugging'' or ``handshaking'', which enhances the mAP at a 50\% IoU threshold.}\

These findings can also be observed for the results on the BABEL benchmark for all considered IoU thresholds (10\%, 30\% and 50\%) as detailed in ~\cref{tab:babel_seg}. Our method significantly outperforms the state-of-the-art results, yielding a +15\% improvement across all subsets at 10\% IoU and a +10\% increase at 30 \%IoU. Furthermore, the proposed approach also shows superior temporal consistency compared to existing methods, as it can be seen in the detection of ``walk'' actions in the sequences shown in \cref{fig:qualit_examples}. Please see the ``Supplementary Material'' for additional qualitative analysis.

\begin{table}[t]
    \caption{Detection mAP@IoU on PKU-MMDv1 cross-subject and cross-view splits.}
    \centering
    \label{tab:pku-mmd-results}
    \resizebox{0.8\linewidth}{!}{%
        \centering
        \begin{tabular}{lcccc}
            \toprule
            \multirow{2}{*}{\textbf{Methods}} & \multicolumn{2}{c}{\textbf{CS}} & \multicolumn{2}{c}{\textbf{CV}} \\
            & 10(\%) & 50(\%) & 10(\%) & 50(\%) \\
            \midrule
            \midrule
            JCRRNN\cite{li2016online} (ECCV'16) & 45.2 & 32.5 & -- & --\\
            Conv. Skel.~\cite{liu2017pku} (MMW'17) & 64.9 & 19.9 & 74.2 & 30.6 \\
            Skel. boxes~\cite{li2017skeleton} (ICMEW'17) & 61.3 & 54.8 & \underline{94.5} & \underline{94.2}\\
            Hi-TRS~\cite{chen2022hitrs} (ECCV'22) & -- & 67.3 & -- & --\\
            LAC (scratch) ~\cite{yang2023lac} (ICCV'23) & \underline{86.5} & 74.6 & 92.9 & 90.1 \\
            USDRL~\cite{wengUSDRL25} (AAAI'25) & -- & \underline{75.7} & -- & --  \\
            Ours & \textbf{93.7} & \textbf{80.9} & \textbf{96.1} & \textbf{94.5}\\
            \bottomrule
        \end{tabular}
    }\vspace{-0.2cm}
\end{table}

\begin{table}[t]
    \caption{Detection mAP@IoU on BABEL subsets~\cite{Yu_Fujiwara_2023}.}
        \label{tab:babel_seg}
        \resizebox{\linewidth}{!}{%
        \centering
            \begin{tabular}{lccccccccc}
            \toprule
            Method & \multicolumn{3}{c}{Subset-1} &  \multicolumn{3}{c}{Subset-2} & \multicolumn{3}{c}{Subset-3} \\
            \midrule
             & 10\% & 30\% & 50\% & 10\% & 30\% & 50\% & 10\% & 30\% & 50\%\\
            \midrule
            \midrule
            ASOT~\cite{xu2024temporally} (CVPR'24) & 42.3 & 34.1 & \underline{24.5} & 40.3 & 33.4 & 23.4 & 27.4 & 21.6 & 14.3\\
            SMQ~\cite{gokay2025skeleton} (ICCV'25) & 40.9 & 32.8 & 22.3 & 43.8 & 37.4 & \underline{27.4} & {38.0} & 29.3 & 19.3\\
            S-WTAL~\cite{Yu_Fujiwara_2023} (AAAI'23) & {48.7} & {39.8} & 21.7 & {61.0} & \underline{50.3} & 19.6 & 35.8 & {31.5} & \underline{20.4} \\
            LAC~\cite{yang2023lac} (ICCV'23) & \underline{55.7} & \underline{46.1} & 24.2 & \underline{63.9} & 48.5 & 26.7 & \underline{59.1} & \underline{48.3} & 20.2 \\
            Ours & \textbf{70.8} & \textbf{58.8} & \textbf{28.1} & \textbf{84.0} & \textbf{67.5} & \textbf{36.1} & \textbf{78.6} & \textbf{72.1} & \textbf{54.5}\\
            \bottomrule
            \end{tabular}
        }
      
\end{table}

\subsection{Ablation and Sensitivity Analysis}
In designing our framework, we evaluated several spatial window encoders~\cite{Shi2sAGCN,yang2021unik,liu2025revealing} commonly used in temporal action detection settings~\cite{yang2023lac}. \Cref{tab:ablation} reports the different combinations of window encoders (stage 1) and temporal encoders (stage 2) we investigated, along with the resulting performance on BABEL. Our approach consistently achieves mAP scores exceeding all evaluated window encoders, demonstrating that the proposed temporal encoder generalizes well across different backbone architectures. Furthermore, the designed lightweight backbone (SWGCN), exhibits significantly faster inference compared to these alternative encoders, achieving nearly twice the speed of PROTOGCN~\cite{liu2025revealing}. 

The designed multi-scale temporal encoder (HydraView) also presents improved detection results when compared to the multi-scale strategies MS-TEMBA~\cite{sinha2025ms} and PDAN~\cite{dai2021pdan}. In addition, HydraView maintains competitive computational efficiency, processing up to 205 sequences per second. In this setting, each sequence consists of features extracted from 2,500 temporal windows that are subsequently refined by the temporal encoder.
Furthermore, we analyze the behavior of our framework under varying numbers of generated virtual viewpoints at training time. This study reveals a consistent performance gain when multiple views are used compared to a single-view setting. Detection accuracy continues to improve as the number of views increases and begins to saturate when more than six viewpoints are incorporated.

\begin{table}[t]
    \caption{{Ablation and sensitivity on BABEL mAP@ 50\%.}}
        \label{tab:ablation}
        \resizebox{\linewidth}{!}{%
        \centering
            \begin{tabular}{ccccccc}
            \toprule
            Window & Temporal &
           \# views & \multirow{2}{*}{Subset-1} &  \multirow{2}{*}{Subset-2} & \multirow{2}{*}{Subset-3} & \multirow{2}{*}{} \\
            Encoder & Encoder & at train & & & \\
            \midrule
            \multicolumn{6}{l}{Different window encoders} & \textbf{(windows/sec)} \\
            \hdashline
            PROTOGCN~\cite{liu2025revealing} & \multirow{3}{*}{HydraView (Ours)} & \multirow{3}{*}{12} & 27.1 & 34.6 & 52.9 & 49\\
            UNIK~\cite{yang2021unik} & & & 27.5 & 35.6 & 53.3 & 77 \\
            SWGCN (Ours) & & & 28.1 & 36.1 & 54.5 & 115 \\
            \midrule
            \multicolumn{6}{l}{Different temporal encoders} & \textbf{(sequences/sec)}\\
            \hdashline
            \multirow{4}{*}{SWGCN (Ours)} & -- & \multirow{4}{*}{12} & 22.3 & 24.9 & 30.1 & -- \\
            & PDAN~\cite{dai2021pdan} & & 24.1 & 25.2 & 32.5 & 187 \\
            & MS-TEMBA~\cite{sinha2025ms} & & 26.5 & 27.5 & 45.8 & 134 \\
            & HydraView (Ours) & & 28.1 & 36.1 & 54.5 & 205 \\
            \midrule
            \multicolumn{6}{l}{Different number of views} & \textbf{(sequences/sec)}\\
            \hdashline
            \multirow{4}{*}{SWGCN (Ours)} & \multirow{4}{*}{HydraView (Ours)} & 1 & 24.7 & 30.2 & 48.2 & \multirow{4}{*}{205} \\
            & & 3 & 27.3 & 35.6 & 53.5 & \\
            & & 6 & 27.9 & 35.9 & 54.1 & \\
            & & 12 & 28.1 & 36.1 & 54.5 & \\
            \bottomrule
            \end{tabular}
        }
\end{table}

\vspace{-0.3cm}

\section{Conclusion}

This paper introduces a novel temporal action detection framework that jointly improves view invariance and temporal consistency. While existing video-based approaches generally lack robustness to viewpoint variations, motion-based detection methods often fail to model temporal relationships across adjacent windows. To overcome these limitations, we design a method based on two main components (SWGCN and HydraView) that leverages multiple viewpoints during training to enhance frame-level action localization in single-view untrimmed videos during inference. Our method combines multi-view learning on short temporal windows to enforce view-invariant yet discriminative motion representations with a multi-view, multi-scale temporal encoder based on recent state space models. The proposed approach achieves improvements over recent state-of-the-art temporal action detection methods on two widely adopted datasets PKU-MMD and BABEL.

\vspace{0.1cm}
\noindent\textbf{Acknowledgements.} This work was funded by TEB Group - Prynel SAS, and by grants ANER MOVIS from ``Conseil Regional de Bourgogne-Franche-Comte'' and ANR MANYVIS (ANR-23-CE23-0003-01), to whom we are grateful.

\clearpage

{\small
\bibliographystyle{IEEE}
\bibliography{
bibs/TAD,
bibs/Datasets,
bibs/AR,
bibs/AR_MV,
bibs/3D_shape,
bibs/Mamba,
bibs/3D_object_rec}

@inproceedings{Yan2018SpatialTG,
  title={Spatial Temporal Graph Convolutional Networks for Skeleton-Based Action Recognition},
  author={Sijie Yan and Yuanjun Xiong and Dahua Lin},
  booktitle={AAAI},
  year={2018}
}

@INPROCEEDINGS{Shi2sAGCN,
  author={Shi, Lei and Zhang, Yifan and Cheng, Jian and Lu, Hanqing},
  booktitle={CVPR}, 
  title={Two-Stream Adaptive Graph Convolutional Networks for Skeleton-Based Action Recognition}, 
  year={2019},
  volume={},
  number={},
  pages={12018-12027},
  doi={10.1109/CVPR.2019.01230}
}

@article{yang2021unik,
      title={UNIK: A Unified Framework for Real-world Skeleton-based Action Recognition}, 
      author={Di Yang and Yaohui Wang and Antitza Dantcheva and Lorenzo Garattoni and Gianpiero Francesca and Francois Bremond},
      year={2021},
      journal={BMVC}
}

@inproceedings{liu2025revealing,
  title={Revealing key details to see differences: A novel prototypical perspective for skeleton-based action recognition},
  author={Liu, Hongda and Liu, Yunfan and Ren, Min and Wang, Hao and Wang, Yunlong and Sun, Zhenan},
  booktitle={CVPR},
  pages={29248--29257},
  year={2025}
}

@inproceedings{BABEL:CVPR:2021,
  title = {{BABEL}: Bodies, Action and Behavior with English Labels},
  author = {Punnakkal, Abhinanda R. and Chandrasekaran, Arjun and Athanasiou, Nikos and Quiros-Ramirez, Alejandra and Black, Michael J.},
  booktitle = {CVPR},
  pages = {722--731},
  month = jun,
  year = {2021},
  doi = {},
  month_numeric = {6}
}

@conference{AMASS:ICCV:2019,
  title = {{AMASS}: Archive of Motion Capture as Surface Shapes},
  author = {Mahmood, Naureen and Ghorbani, Nima and Troje, Nikolaus F. and Pons-Moll, Gerard and Black, Michael J.},
  booktitle = {ICCV},
  pages = {5442--5451},
  month = oct,
  year = {2019},
  month_numeric = {10}
}

@article{liu2017pku, 
  title={PKU-MMD: A Large Scale Benchmark for Continuous Multi-Modal Human Action Understanding},
  author={Chunhui, Liu and Yueyu, Hu and Yanghao, Li and Sijie, Song and Jiaying, Liu},
  journal={ACM Multimedia workshop},
  year={2017}
}

@article{Dai:TPAMI:2023,
  author={Dai, Rui and Das, Srijan and Sharma, Saurav and Minciullo, Luca and Garattoni, Lorenzo and Bremond, Francois and Francesca, Gianpiero},
  journal={TPAMI}, 
  title={Toyota Smarthome Untrimmed: Real-World Untrimmed Videos for Activity Detection}, 
  year={2023},
  volume={45},
  number={2},
  pages={2533-2550},
  doi={10.1109/TPAMI.2022.3169976}}

@inproceedings{gu2024mamba,
  title={Mamba: Linear-time sequence modeling with selective state spaces},
  author={Gu, Albert and Dao, Tri},
  booktitle={First conference on language modeling},
  year={2024}
}

@inproceedings{zhu2024vision,
  title={Vision mamba: Efficient visual representation learning with bidirectional state space model},
  author={Zhu, Lianghui and Liao, Bencheng and Zhang, Qian and Wang, Xinlong and Liu, Wenyu and Wang, Xinggang},
  journal={arXiv preprint arXiv:2401.09417},
  year={2024},
booktitle = {ICML}
}

@article{liu2024vmamba,
  title={Vmamba: Visual state space model},
  author={Liu, Yue and Tian, Yunjie and Zhao, Yuzhong and Yu, Hongtian and Xie, Lingxi and Wang, Yaowei and Ye, Qixiang and Jiao, Jianbin and Liu, Yunfan},
  journal={NeurIPS},
  volume={37},
  pages={103031--103063},
  year={2024}
}

@inproceedings{lu2025jamma,
  title={Jamma: Ultra-lightweight local feature matching with joint mamba},
  author={Lu, Xiaoyong and Du, Songlin},
  booktitle={CVPR},
  year={2025}
}

@inproceedings{lea2017temporal,
  title={Temporal convolutional networks for action segmentation and detection},
  author={Lea, Colin and Flynn, Michael D and Vidal, Rene and Reiter, Austin and Hager, Gregory D},
  booktitle={CVPR},
  pages={156--165},
  year={2017}
}

@article{wang2022empirical,
  title={An empirical study on temporal modeling for online action detection},
  author={Wang, Wen and Peng, Xiaojiang and Qiao, Yu and Cheng, Jian},
  journal={CISIS},
  volume={8},
  pages={1803--1817},
  year={2022},
  publisher={Springer}
}

@inproceedings{li2016online,
  title={Online human action detection using joint classification-regression recurrent neural networks},
  author={Li, Yanghao and Lan, Cuiling and Xing, Junliang and Zeng, Wenjun and Yuan, Chunfeng and Liu, Jiaying},
  booktitle={ECCV},
  pages={203--220},
  year={2016},
  organization={Springer}
}

@inproceedings{li2017skeleton,
  title={Skeleton boxes: Solving skeleton based action detection with a single deep convolutional neural network},
  author={Li, Bo and Chen, Huahui and Chen, Yucheng and Dai, Yuchao and He, Mingyi},
  booktitle={ICMEW},
  pages={613--616},
  year={2017},
  organization={IEEE}
}

@inproceedings{dai2021pdan,
  title={Pdan: Pyramid dilated attention network for action detection},
  author={Dai, Rui and Das, Srijan and Minciullo, Luca and Garattoni, Lorenzo and Francesca, Gianpiero and Bremond, Fran{\c{c}}ois},
  booktitle={WACV},
  year={2021}
}

@InProceedings{chen2022hitrs,
author="Chen, Yuxiao
and Zhao, Long
and Yuan, Jianbo
and Tian, Yu
and Xia, Zhaoyang
and Geng, Shijie
and Han, Ligong
and Metaxas, Dimitris N.",
title="Hierarchically Self-supervised Transformer for Human Skeleton Representation Learning",
booktitle="ECCV",
year="2022"
}

@inproceedings{dai2022ms,
  title={Ms-tct: Multi-scale temporal convtransformer for action detection},
  author={Dai, Rui and Das, Srijan and Kahatapitiya, Kumara and Ryoo, Michael S and Br{\'e}mond, Fran{\c{c}}ois},
  booktitle={CVPR},
  year={2022}
}

@article{Yu_Fujiwara_2023,
title={Frame-Level Label Refinement for Skeleton-Based Weakly-Supervised Action Recognition},
volume={37},
DOI={10.1609/aaai.v37i3.25439},
number={3}, 
journal={AAAI}, 
author={Yu, Qing and Fujiwara, Kent}, 
year={2023}, 
month={Jun.}, 
pages={3322-3330} 
}

@inproceedings{yang2023lac,
  title={Lac-latent action composition for skeleton-based action segmentation},
  author={Yang, Di and Wang, Yaohui and Dantcheva, Antitza and Kong, Quan and Garattoni, Lorenzo and Francesca, Gianpiero and Bremond, Francois},
  booktitle={ICCV},
  pages={13679--13690},
  year={2023}
}

@article{liu2024harnessing,
  title={Harnessing Temporal Causality for Advanced Temporal Action Detection},
  author={Liu, Shuming and Sui, Lin and Zhang, Chen-Lin and Mu, Fangzhou and Zhao, Chen and Ghanem, Bernard},
  journal={arXiv preprint arXiv:2407.17792},
  year={2024}
}

@article{chen2024video,
  title={Video mamba suite: State space model as a versatile alternative for video understanding},
  author={Chen, Guo and Huang, Yifei and Xu, Jilan and Pei, Baoqi and Chen, Zhe and Li, Zhiqi and Wang, Jiahao and Li, Kunchang and Lu, Tong and Wang, Limin},
  journal={IJCV},
  volume={134},
  number={20}, 
  year={2026}
}

@inproceedings{xu2024temporally,
  title={Temporally consistent unbalanced optimal transport for unsupervised action segmentation},
  author={Xu, Ming and Gould, Stephen},
  booktitle={CVPR},
  pages={14618--14627},
  year={2024}
}

@inproceedings{zhu2024dual,
  title={Dual detrs for multi-label temporal action detection},
  author={Zhu, Yuhan and Zhang, Guozhen and Tan, Jing and Wu, Gangshan and Wang, Limin},
  booktitle={CVPR},
  pages={18559--18569},
  year={2024}
}

@inproceedings{tian2025stitch,
  title={Stitch, contrast, and segment: Learning a human action segmentation model using trimmed skeleton videos},
  author={Tian, Haitao and Payeur, Pierre},
  booktitle={AAAI},
  volume={39},
  year={2025}
}

@article{sinha2025ms,
  title={MS-Temba: Multi-scale temporal mamba for efficient temporal action detection},
  author={Sinha, Arkaprava and Raj, Monish Soundar and Wang, Pu and Helmy, Ahmed and Das, Srijan},
  journal={CVPR},
  year={2026}
}

@inproceedings{gokay2025skeleton,
  title={Skeleton Motion Words for Unsupervised Skeleton-Based Temporal Action Segmentation},
  author={G{\"o}kay, Uzay and Spurio, Federico and Bach, Dominik R and Gall, Juergen},
  booktitle={ICCV},
  year={2025}
}

@inproceedings{wengUSDRL25,
  title={USDRL: Unified Skeleton-Based Dense Representation Learning with Multi-Grained Feature Decorrelation},
  author={Wanjiang Weng and Hongsong Wang and Junbo Wang and Lei He and Guosen Xie},
  booktitle={AAAI},
  year={2025}
}

@inproceedings{tian2025duoclr,
  title={DuoCLR: Dual-Surrogate Contrastive Learning for Skeleton-based Human Action Segmentation},
  author={Tian, Haitao},
  booktitle={ICCV},
  year={2025}
}
}

\clearpage
\appendix

\twocolumn[{%
 \centering
 \Large \textbf{[Supplementary Material] -- \\ Improving Viewpoint-Invariance and Temporal Consistency for Action Detection} \\[1.5em]
}]

In this supplementary material, we provide additional implementation details and experimental results to the main paper. \cref{sec:imp_details} mainly discusses the rationale behind the SWGCN design and highlights its advantages over the baseline ST-GCN. \cref{sec:fur_analysis} presents additional quantitative and qualitative experiments, specifically evaluating the model’s performance with multi-view inputs at inference time.

\section{Implementation Details}
\label{sec:imp_details}

\subsection{SWGCN}

The proposed spatio-temporal backbone departs from the original ST-GCN~\cite{Yan2018SpatialTG} design in several key aspects to better accommodate short-window Temporal Action Detection (TAD), as listed in \cref{tab:stgcn_tad_comparison}. ST-GCN was originally introduced for clip-level action classification and employs temporal downsampling and fixed kernel configurations to progressively increase the temporal receptive field over long sequences. While effective for classification, such temporal downsampling reduces frame-level resolution and can obscure fine-grained temporal boundaries, which are critical for precise action localization. In contrast, we remove temporal downsampling entirely, ensuring that the temporal resolution of the feature maps matches that of the input window, thereby preserving subtle motion cues and boundary information within short clips.

Furthermore, ST-GCN adopts a uniform temporal convolution kernel size across layers, implicitly assuming a fixed temporal scale for action modeling. This assumption is less suitable for TAD, where the relevant temporal context is strongly constrained by the window length~\cite{lea2017temporal}. We therefore adjust the temporal kernel size to explicitly control the temporal receptive field, aligning it with the short temporal extent of the input (e.g., 16 frames). This avoids excessive temporal smoothing and prevents the receptive field from exceeding the effective temporal support of the window, which would otherwise dilute discriminative motion patterns.

Finally, ST-GCN relies on Batch Normalization, which assumes sufficiently large and homogeneous mini-batches. In TAD settings, training is often performed with small batch sizes due to high temporal resolution and sliding-window sampling, making batch statistics noisy and unstable. To address this limitation, we replace Batch Normalization with Group Normalization, which normalizes features independently of the batch dimension. This modification leads to more stable optimization and improved feature consistency across windows, a property that is particularly important for dense temporal prediction tasks such as TAD.

\vspace{-0.4cm}

\begin{table}[t]
\centering
\caption{Comparison between the standard ST-GCN architecture and the proposed TAD-oriented spatio-temporal backbone (SWGCN).}  \vspace{0.2cm}
\label{tab:stgcn_tad_comparison}
\resizebox{\linewidth}{!}{%
    \begin{tabular}{lcc}
    \hline
    \textbf{Component} & \textbf{ST-GCN} & \textbf{SWGCN} \\
    \hline
    Target task & Action classification & Temporal action detection \\
    Input length & Long clips (e.g., 64--300) & Short windows (e.g., 16) \\
    Temporal downsampling & Yes (stride $>1$) & No (stride $=1$) \\
    Temporal kernel size & 9 & Adjusted to window length  \\
    Temporal receptive field & Progressively expanded & Explicitly controlled \\
    Normalization & Batch Normalization & Group Normalization \\
    \hline
    \end{tabular}
}
\end{table}

\begin{table}[t]
    \caption{{Inference with mutiple views for inference on BABEL mAP@ 50\%.}}
    \vspace{0.2cm}
        \label{tab:mv-infer}
        \resizebox{\linewidth}{!}{%
        \centering
            \begin{tabular}{cccccc}
            \toprule
            \multirow{2}{*}{Method} &
           \# views & \multirow{2}{*}{Subset-1} &  \multirow{2}{*}{Subset-2} & \multirow{2}{*}{Subset-3} & \multirow{2}{*}{} \\
            & at infer. & & & \\
            \midrule
            \multicolumn{5}{l}{Different number of views} & \textbf{(sequences/sec)}\\
            \hdashline
            \multirow{4}{*}{Ours} & 1 & 28.1 & 36.1 & 54.5 & 200 \\
            & 3 & 28.5 & 37.3 & 55.7 & 153 \\
            & 6 & 29.1 & 39.4 & 57.2 & 97 \\
            & 12 & 30.1 & 40.7 & 57.6 & 50 \\
            \bottomrule
            \end{tabular}
        }
\end{table}

\begin{figure*}[t]
\centering
\includegraphics[width=\linewidth]{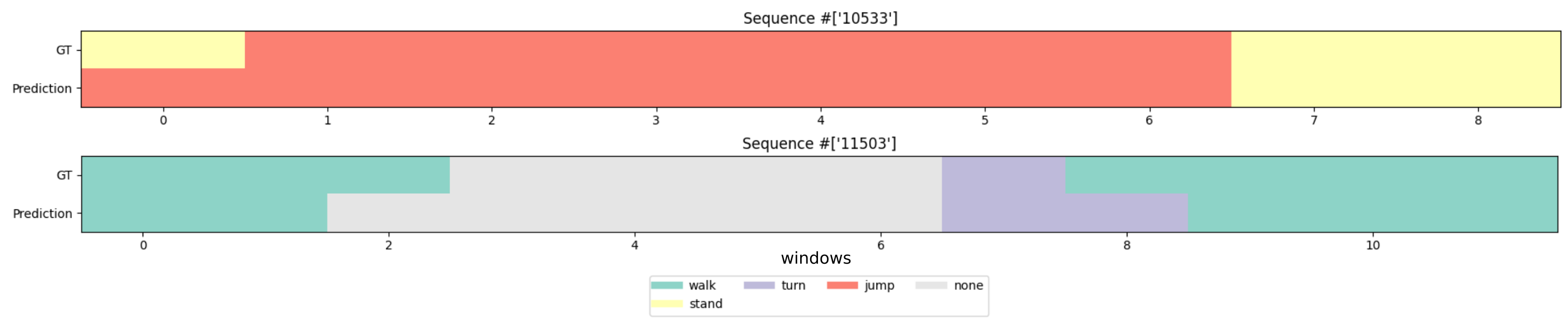}
\caption{\textbf{Results on two sequences from BABEL (split 1).} The predictions indicate the classification for each temporal window. Each action instance is correctly labeled. While some Detections exhibit minor temporal shifts at the boundaries, they maintain an Intersection over Union (IoU) exceeding 50\%.}
\label{fig:qualit_babel}
\end{figure*}

\begin{figure*}[t]
\centering
\includegraphics[width=\linewidth]{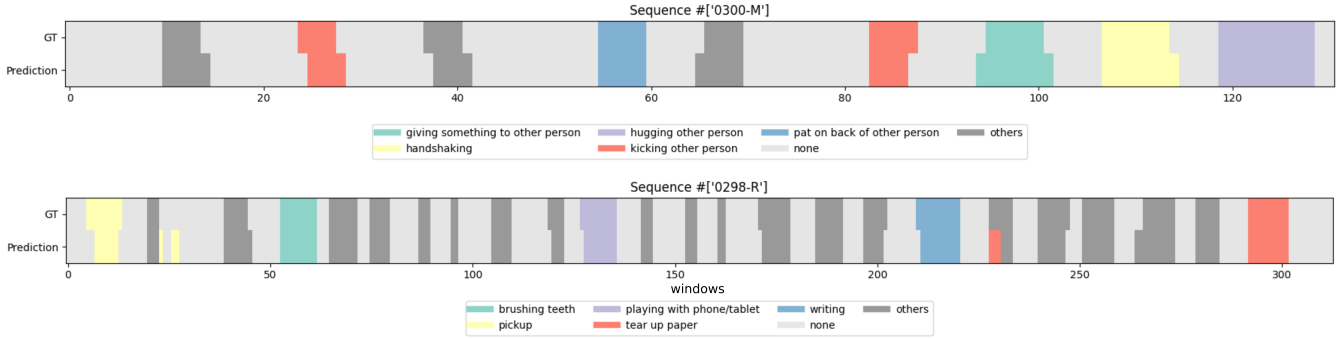}
\caption{\textbf{Results on two sequences from PKUMMDv1 (cross-subject split).} Our model effectively scales to long-form sequences (up to 4 minutes) involving a large action vocabulary (51 classes). For visual clarity, we aggregate ``others'' instances into a single category, though they retain their individual class labels for evaluation purposes. The results demonstrate the model's ability to maintain temporal consistency across extended durations.}
\label{fig:qualit_pkummd}
\end{figure*}

\subsection{Occlusion Estimations}

{The occlusion estimation module, introduced in Sec. 3.1, simulates the 2D skeleton artifacts typical of varied camera perspectives. In real-world scenarios, joints positioned on the far side of the actor relative to the sensor often yield low-confidence scores and are subsequently filtered. To model this phenomenon, we approximate the human torso as a 3D reference plane. For each joint, we compute the intersection between this plane and the vector extending from the camera to the joint using standard line-plane intersection equations. A joint is flagged as occluded if this vector intersects the torso plane, effectively mimicking the self-occlusion observed in monocular 2D pose estimation.}

\subsection{Fuser Component}

{The Multi-Scale Fuser is centered around a VisionMamba layer~\cite{zhu2024vision}, chosen for its linear scaling and global receptive field. To harmonize features across different granularities, multi-scale inputs are first projected through a dense layer and stabilized via layer normalization. These refined features are then processed by the VisionMamba backbone with the final predictions generated by a dense MLP head.}

\section{Further Analysis}
\label{sec:fur_analysis}
\subsection{Multi-view Inference}
We also evaluate the performance of our end-to-end architecture across varying numbers of simultaneous views taken into account at inference as well (\cref{tab:mv-infer}). This scenario is relevant if multiple video streams would be available such as on surveillance, CCTV or with camera rigs. Please notice however that all the previous results has been done considering a single view at inference for a fair comparison to the baselines (these results corresponds to 1 view row). Adding extra views at inference time brings supplementary information, therefore increasing the network accuracy. This demonstrates the ability of our network to aggregate features coming from different views at inference time as well. Even when considering 12 video streams at inference, our network achieves 50 sequences per second, contrasting favorably with existing temporal encoders that exhibit linear scaling complexity relative to the number of views. In this context, a single sequence comprises 2,500 temporal windows distributed across 12 views.

\subsection{Qualitative Results}
This section provides additional qualitative visualizations to further evaluate the proposed framework. \cref{fig:qualit_babel} illustrates the predicted action for each temporal window throughout the sequence. The observed temporal shifts between the ground truth (GT) and our predictions often stem from annotation ambiguity. Boundaries may vary based on whether an action is defined by the initial perceived motion or the physical displacement of joints. We also present qualitative results on the PKUMMDv1 dataset in \cref{fig:qualit_pkummd}. These longer sequences pose a significant challenge due to the density of action instances and the complex temporal dependencies between them. While our model accurately identifies most action boundaries, minor false detections remain, such as the ``pickup'' and ``tear-up paper'' instances in the bottom sequence, highlighting areas for future refinement.

\subsection{Per-Class Analysis}

{The per-class performance across the three BABEL splits demonstrates the model's robust generalization capabilities, maintaining high Mean Average Precision (mAP) regardless of the specific data partitioning. Notably, the model excels at identifying structured, periodic actions such as ``dance'' (96\%) and ``sit'' (94\%), suggesting that the temporal features of these classes are highly discriminative.

Conversely, more nuanced or transitional actions like ``turn'' (53\%) and ``stand'' (68\%) show relatively lower mAP scores. This likely stems from the inherent ambiguity in their start and end points, which pose a greater challenge for precise temporal localization (as observed in \cref{fig:qualit_babel}). Despite these variations, the results confirm that the model avoids complete misdetection of any single category, maintaining a performance floor above 50\% across all evaluated splits. This consistency validates that the architecture is effectively capturing universal motion primitives rather than over-fitting to specific split distributions.}

\begin{table}[t]
    \caption{{Per-class analysis of our method on the different splits of BABEL, with an IoU of 10\%.}}
        \label{tab:per-class}
        \resizebox{\linewidth}{!}{%
        \begin{tabular}{cccccc}
        \toprule
        \centering
        BABEL split1 & mAP & BABEL split2 & mAP & BABEL split3 & mAP \\
        %split1 & mAP & split2 & mAP  & split3 & mAP\\
        \midrule
        walk & 84 & sit & 94 & jog & 79\\
        stand & 68 & run & 77 & wave & 68\\
        turn & 53 & stand-up & 89 & dance & 96\\
        jump & 79 & kick & 76 & gesture & 74\\
        \bottomrule
        \end{tabular}
        }
\end{table}

\end{document}